\documentclass{article}

\usepackage{microtype}
\usepackage{graphicx}
\usepackage{subfigure}
\usepackage{booktabs} 
\usepackage[inline]{enumitem}

\usepackage[hidelinks]{hyperref}

\usepackage[accepted]{icml2024_adapted_for_arxiv}

\usepackage{amsmath}
\usepackage{amssymb}
\usepackage{mathtools}
\usepackage{amsthm}
\usepackage{enumitem}
\usepackage[capitalize,noabbrev]{cleveref}
\usepackage{xurl} 
\usepackage[subtle]{savetrees}

\theoremstyle{plain}

\theoremstyle{definition}

\theoremstyle{remark}

\usepackage[textsize=tiny]{todonotes}

\icmltitlerunning{Mission Critical -- 
Satellite Data is a Distinct Modality in Machine Learning}

\begin{document}

\title{Mission Critical -- 
Satellite Data is a Distinct Modality in Machine Learning}
\twocolumn[
\icmltitle{Mission Critical -- 
Satellite Data is a Distinct Modality in Machine Learning
}




\icmlsetsymbol{equal}{*}
\begin{icmlauthorlist}
\icmlauthor{Esther Rolf}{equal,harvard,cu}
\icmlauthor{Konstantin Klemmer}{msr}
\icmlauthor{Caleb Robinson}{msft}
\icmlauthor{Hannah Kerner}{equal,asu}

\end{icmlauthorlist}
\icmlaffiliation{harvard}{Harvard Data Science Initiative and Center for Research on Computation and Society, Harvard University}
\icmlaffiliation{asu}{School of Computing and Augmented Intelligence, Arizona State University}
\icmlaffiliation{msr}{Microsoft Research}
\icmlaffiliation{msft}{Microsoft AI for Good Research Lab}
\icmlaffiliation{cu}{University of Colorado, Boulder}
\icmlcorrespondingauthor{Hannah Kerner}{hkerner@asu.edu}
\icmlcorrespondingauthor{Esther Rolf}{esther.rolf@colorado.edu}

 \vskip 0.3in
 ]



\printAffiliationsAndNotice{$^{\ast}$Hannah Kerner and Esther Rolf jointly conceived and led this position paper, with equal contributions.} 

\begin{abstract}
Satellite data has the potential to inspire a seismic shift for machine learning---one in which we rethink existing practices designed for traditional data modalities. 
As machine learning for satellite data (SatML) gains traction for its real-world impact, our field is at a crossroads. We can either continue applying ill-suited approaches, or we can
initiate a new research agenda that centers around the unique characteristics and challenges of satellite data.
This position paper argues that satellite data constitutes a distinct modality for machine learning research and that we must recognize it as such to advance the quality and impact of SatML research across theory, methods, and deployment. We outline 
critical discussion questions and actionable suggestions to transform SatML from merely an intriguing application area to a dedicated research discipline that helps move the needle on big challenges for machine learning and society.

\end{abstract}



\begin{figure*}[thb!]
    \centering
    \includegraphics[width=.98\textwidth]{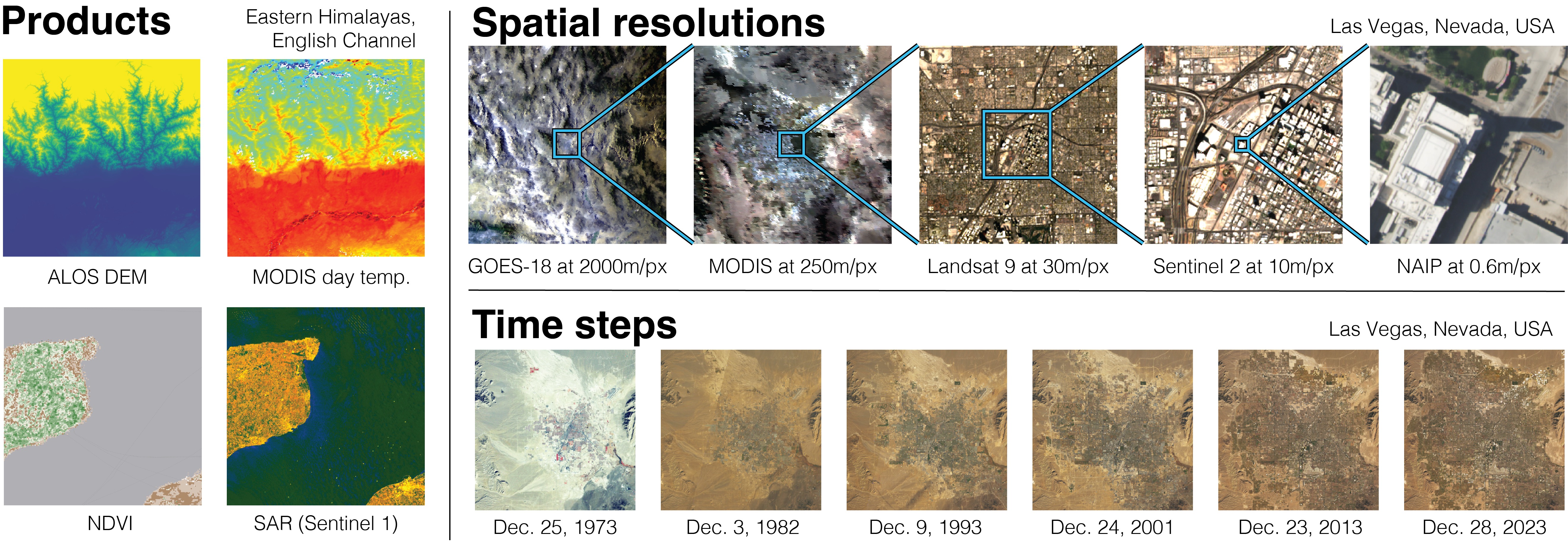}
    \caption{\textbf{Satellite images of the same location can vary widely} depending on factors like spatial resolution and cropping extent,  temporal dimension, and satellite mission or instrument. ML methods that leverage these factors can drastically outperform methods for general images.}
    \label{fig:multi-sat-example}
\end{figure*}

\section{Introduction}
Hundreds of remote sensing satellites continuously monitor the Earth's surface, creating petabyte-scale time series datasets. These datasets hold information needed to measure and address pressing planetary-scale challenges including climate change~\cite{yang2013role}, poverty~\cite{jean2016combining}, food insecurity~\cite{nakalembe2020urgent}, biodiversity loss~\cite{skidmore2021priority}, and many other global goals~\cite{kavvada2020towards}. 
However, 
extracting actionable insights from satellite data at scale requires specialized analysis.

Machine learning, especially deep learning, was born to extract insights from raw data at scale. However, current ML solutions designed for other data modalities, such as natural images or language, are sub-optimal for satellite data. 

Satellite data presents challenges
and opportunities distinct from other data modalities (\cref{fig:multi-sat-example}). Unlike natural images, the size of targets in satellite images span a logarithmic scale from $<1$m (e.g., trees) to $>1$km (e.g., forests). Temporal patterns in satellite time series also span logarithmic scales, from hours or days (e.g., floods) to years or decades (e.g., sea level rise). 
Data are acquired using a variety of sensors that capture diverse spectral channels (beyond 3-channel RGB) and precise measurements (beyond 8 bits).
Satellites collect data over the entire surface of the Earth at fixed time intervals and spatial resolutions. Observations are acquired from an overhead perspective from fixed altitudes and lack a ``natural'' orientation, unlike natural images. 
  
While there has been increasing interest in ML for satellite data (SatML) (\citet{zhu2017deep}; Table A\ref{tab:workshops}), SatML research as a whole falls short on these challenges and opportunities. The mainstream approach to SatML has been to adapt or ``lift and shift'' solutions designed for other modalities, especially natural images, to satellite data with minimal tailoring.
Rather than tackling novel or outstanding challenges, many studies propose ML solutions to well-resolved or low-priority problems in the field of remote sensing~\cite{tuia2021toward}. Researchers favor large, ``state of the art'' models designed for other modalities without regard for the unique characteristics and deployment considerations of SatML models. This comes at a cost of performance, scalability, and missed research opportunities.

The unique properties of satellite data and the challenges of SatML---as with other data modalities such as natural images and text---warrant specialized methods for collecting, modelling, and interpreting data. 
For example, self-supervised learning techniques have been proposed to explicitly learn spatiotemporal relationships in geospatial data, such as nearby locations being more similar than distant locations~\cite{jean2019tile2vec} or the same location varying in appearance over different seasons~\cite{manas2021seasonal}. \citet{marcos2018land} proposed a small rotation-equivariant neural network that outperformed a larger convolutional neural network in SatML land cover mapping tasks. 
While these examples illustrate the potential for specialized SatML approaches, we need to encourage more such research by placing greater value as a community on efforts that tackle the distinct challenges and opportunities of satellite data.

In this position paper, we argue that \textbf{satellite data constitutes a distinct data modality for ML} 
and that
\textbf{we must recognize it as such to advance the quality and impact of SatML research.}
Approaches designed for other ML data modalities are ill-fitted for the distinct challenges and opportunities of modeling satellite data and deploying SatML systems (\S2). We must prioritize the development of specialized SatML approaches rather than limiting ourselves to ``lifting and shifting'' existing approaches (\S3). Leveraging satellite data can also enrich cross-cutting ML research topics such as distribution shift, self-supervised learning, and multi-modal learning (\S4). 
%
%
Realizing the full potential of SatML will require substantial shifts in priorities and practices that the ML community must discuss and act on together (\S5).

Prior survey papers detail ML problem settings and applications for remote sensing data~\cite{zhu2017deep}, promising research opportunities at the intersection of Earth observation and ML~\cite{tuia2023artificial,tuia2021toward,reichstein2019deep,bergen2019machine,rolnick2022tackling}, and data-centric ML analyses for remote sensing data~\cite{roscher2023data}.
The goal of this position paper is to present arguments to the general ML community about the opportunities and distinctions of SatML 
within ML at large.
We hope that our arguments provide a path forward -- as well as a warning call -- for researchers who seek to innovate in SatML or leverage satellite data in cross-cutting ML research.

Many of our arguments for satellite data extend naturally to the broader classes of remote sensing and geospatial data (e.g. aerial images, ERA-5 climate data). We focus on here on SatML, with implications for to the connected sub-fields of ML with remote sensing and geospatial ML.  

\section{Satellite data is a distinct modality}
\label{sec:satml_distinct_challenges}

SatML data have distinct characteristics that are not addressed by ML methods designed for other data modalities.

\subsection{Satellite data has unique characteristics} \label{subsec:sat_data_differences}
Satellite data are a type of remotely sensed data encompassing diverse datasets with unique characteristics (\cref{fig:multi-sat-example}). Satellite data are typically stored in a raster format as a tensor with height, width, and channel dimensions. 
Temporal information may be included by stacking spatially-aligned rasters along a fourth dimension. While this representation shares similarities with images and videos, several characteristics  make modeling satellite data fundamentally different.

\paragraph{Spatial and temporal scales}
\textit{SatML can model target phenomena that span logarithmic scales in the spatial and temporal dimensions} (\cref{fig:multi-sat-example}). 
Objects and landforms can range in size from less than one meter (e.g., trees or cars) to many kilometers (e.g., forests or wildfires). Temporal patterns can manifest over hours (e.g., earthquakes), weeks (e.g., new construction), seasons (e.g., crop cultivation), years (e.g., glacial retreat), and decades (e.g., sea level rise). 

Satellite data is acquired from different instruments and orbits that determine the spatial and temporal resolution of a dataset. Spatial resolution describes the size of the area covered by a single pixel on the Earth's surface. Temporal resolution describes the frequency at which observations of the same area are captured. Some satellites capture data on-demand with very high spatial resolution but less frequent temporal coverage. For example, the WorldView-3 satellite captures images with $0.3 \times 0.3$ m/pixel spatial resolution, but each observation covers a small geographic extent and is infrequent in time. At the other extreme, the MODIS satellite captures $500$ $\times$ $500$ m/pixel images daily with a large geographic extent.

\paragraph{Spectral channels}
\textit{SatML can model many spectral channels with greater diversity and precision than standard 8-bit RGB images} (\cref{fig:multi-sat-example}).
Different satellite sensors capture light from different regions of the electromagnetic spectrum. 
Passive sensors like optical and thermal imagers measure light reflected or emitted at different wavelengths from the Earth's surface. Active sensors like synthetic aperture radar (SAR), Lidar, and radar altimetry emit radiation and measure the returned signal. 
For example, Sentinel-1 captures 2-channel (horizontal and vertical polarization) SAR images, while Sentinel-2 captures 13-channel optical images from the visible, near-infrared and shortwave infrared spectrum. 
Data are typically stored in GeoTIFF or other specialized formats that allow higher radiometric resolution than standard 8-bit image formats like JPEG (e.g., Sentinel-2 has 12-bit resolution). 

ML methods developed for 3-channel RGB images are sub-optimal for modeling satellite datasets (\S\ref{sec:tailored}). In addition, common ML libraries lack support for many-channel satellite images. For example, TorchVision and other libraries with pre-trained models assume images are 3-channel and thus cannot be used for transfer learning with most satellite datasets.



\paragraph{Data volume}
\textit{SatML methods must efficiently process enormous, continually growing data volumes.}
Satellite datasets cover the entire globe at high spatial and temporal resolutions. These datasets date back to 1972 and will continue to grow in the future. 
As of January 2024, the European Space Agency's Copernicus Data Space Ecosystem 
contained 66 petabytes of public satellite data from the Sentinel missions~\cite{copernicus_dataspace_dashboard}.
The upcoming NASA-ISRO Synthetic Aperture Radar (NISAR) mission will generate up to 140 petabytes of data (85 TB/day)~\cite{blumenfeld2017getting}.
These petabyte-scale datasets must be processed efficiently at inference time (\S\ref{sec:deployment}) and present data curation challenges at training time.

Satellite data archives are much larger than existing large datasets used in ML research. For example, the text dataset of snapshots from the Common Crawl database used to train GPT-3 was 45 TB (and 570 GB after filtering)~\cite{brown2020language}. 
LAOIN-5B, the largest paired text-image dataset today, is $\approx 220$ TB~\cite{schuhmann2022laion}. The ILSVRC 2012 ImageNet dataset~\cite{imagenet}, widely used for vision model benchmarking and pre-training, is $\approx 150$ GB.

\begin{figure}[tb]
    \centering    \includegraphics[width=\linewidth]{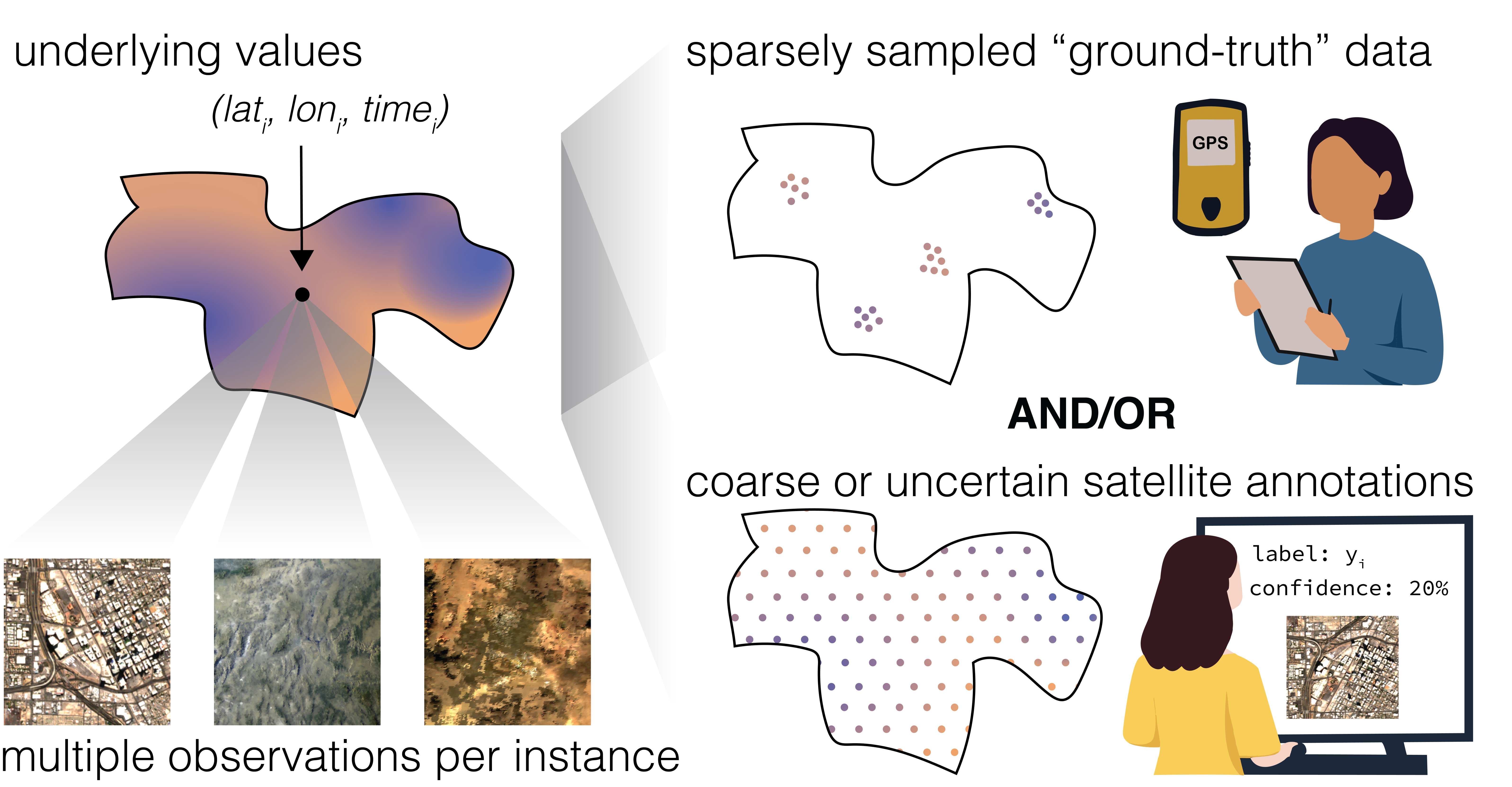}
    \caption{\textbf{In SatML, multiple observations and multiple (or no) labels may correspond to a given (lat, lon, time) index}, whereas in many ML settings, labels are defined directly from images. }
    \label{fig:data_labels}
\end{figure}

\paragraph{Annotations}
\label{sec:label_data}
\textit{SatML must handle small, sparse, and biased labeled datasets that have variable schemas and quality.}
In contrast to many ML settings where observations, $X$, and labels, $y$, are definitive pairs (e.g., $X$ = a photo and $y$ = a label/interpretation of that photo), in SatML, label annotations are often generated irrespective of satellite observations.
Instead, labels can be paired with many different choices of satellite observations corresponding to the label's location and time index (\cref{fig:data_labels}).
``Ground-truth'' labels are obtained by a physical visit to a specific location to obtain an annotation indexed by geographic coordinates and time~\cite{radiantearth-groundtruth}.
These often involve careful sampling informed by spatio-temporal statistics and are generally expensive and difficult to acquire. As a result, ground-truth datasets are typically small, sparse, spatio-temporally clustered, and specialized in a way that can make harmonization with other datasets difficult~\cite{maskey2020advancing}.

Even when labels are created from expert annotations of  satellite observations, severe sampling biases can occur~\cite{reichstein2019deep}. Several large-scale datasets have increased global coverage of annotated image datasets, such as Functional Map of the World~\cite{christie2018functional} and BigEarthNet~\cite{sumbul2019bigearthnet},
but are still more densely sampled in highly populated areas in the Global North.

\subsection{Deployment challenges of dense prediction}
\label{sec:deployment}

For most deployment settings of SatML, the desired output is a set of dense predictions that form a map of an entire geographic region (\cref{fig:spatial_crossval}). For example, crop type classification models generate a predicted map of every pixel in which a specific crop is growing~\cite{nakalembe2023considerations}, and disaster response maps assess damage across every pixel in an affected area~\cite{xu2019building}. 

\textit{The need for dense prediction in deployment makes model efficiency an important SatML consideration.} 
For context, at 10m/pixel resolution, the world's land area is represented by nearly 1.5 trillion pixels. Depending on the task requirements, predictions over such data volumes may need to be updated frequently (e.g., sub-weekly for deforestation monitoring~\cite{diniz2015deter}). Such prediction volumes can be expensive and time-consuming. Model efficiency is thus one of the foremost deployment considerations for end-users~\cite{worldcerealsbenchmark,hengl2017soilgrids250m,robinson2019large}.

Dense prediction also presents opportunities. 
Models can be applied to overlapping extents or fields-of-view, resulting in multiple predictions per pixel that can be averaged for more consistent maps~\cite{huang2018tiling}.
Dense predictions from SatML models can easily be visualized as a larger map, rather than as a set of individual predictions. This can illuminate failure modes for predictive models in a way that is not possible in natural image datasets ~\cite{zvonkov2023openmapflow}. 

\subsection{Evaluation challenges of spatio-temporal ML}
\label{sec:evaluation}

\textit{Traditional ML evaluation practices fall short of the needs for SatML.
} Many recent SatML papers follow a conventional uniform at random assignment of data to training and test sets. 
While there are spatial interpolation tasks where in-sample performance is relevant,
out-of-sample use cases are often the ones where SatML models are most valuable~\cite{rolf2023evaluation}. 
In these settings, uniformly at random sampled test splits can be a serious limitation to understanding model performance. 
For example, 
%
if labels are spatially auto-correlated and clustered, 
assigning data instances to training and evaluation splits uniformly at random will likely overestimate model performance in regions outside of available clusters.
%

\begin{figure}
    \centering
    \includegraphics[width=.85\linewidth]{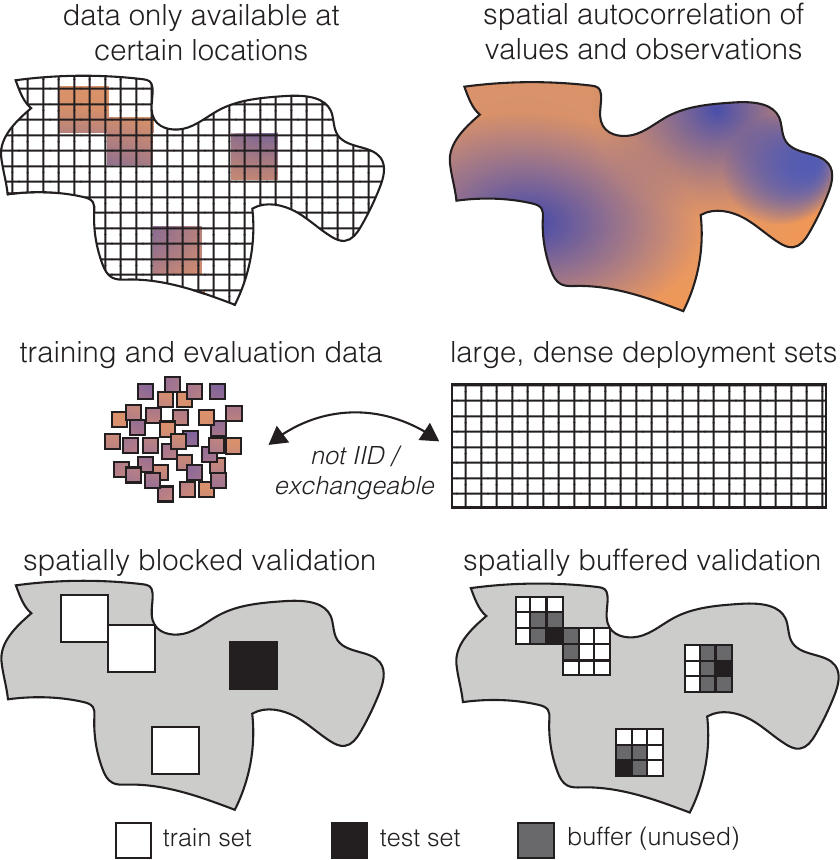}
    \caption{\textbf{SatML has distinct considerations for deployment and evaluation.} 
    Deployment datasets are often dense, and much larger than training datasets.
    Spatio-temporal covariate shifts necessitate spatially aware model validation for out-of-sample model deployment. Figure adapted in part from \citet{zvonkov2023openmapflow}.}
    \label{fig:spatial_crossval}
\end{figure}

%
Spatially aware holdout and cross-validation methods have been designed to
test how models might perform outside the regions in which they were trained. These include blocking or buffering the distance between a train and test set (\cref{fig:spatial_crossval}) or parametrically varying the distance between train and test data~\cite{roberts2017cross, le2014spatial, pohjankukka2017estimating, airola2019spatial,wang2023model}. 

Estimating the actual accuracy of a predicted map requires carefully sampled ground truth data to measure the predictions against~\cite{wadoux2021spatial,stehman2019key}.
Yet, high-quality and uniformly sampled data is scarce (\S\ref{sec:label_data}).
The design and sampling of datasets used to evaluate predictions and derived maps over time and space is thus a major endeavor 
\cite{olofsson2014good,tsendbazar2021towards}. 

Assessing whether published SatML models and maps add value (i.e., predictions are accurate, calibrated, and better than random) remains a significant hurdle to developing reliable SatML for real-world settings.
%
%
SatML inherits many evaluation challenges of geospatial ML, such as (i) reliably estimating accuracy and uncertainty from cluster-sampled reference data ~\cite{mila2022nearest,wang2023model} and (ii) developing notions of  ``regions of applicability" to describe limits on where models are useful~\cite{meyer2022machine}.
Given the distinct deployment (\S\ref{sec:deployment}) and ethical considerations  (\S\ref{sec:ethics}) of SatML models, there is also a need for dedicated performance metrics that measure computational efficiency, privacy, transparency, usability, and fairness. 

\subsection{Distinct ethical concerns of SatML}
\label{sec:ethics}
%
%
%
While many satellite data products have been in the public domain for decades (\S\ref{subsec:sat_data_differences}), SatML systems increase the scale and ease at which people and their activities can be monitored.
In many countries, property laws protect land owners from physical invasion of privacy or unsanctioned use of land by others, but these laws do not clearly extend to satellite representations of the same space.
\textit{People have no clear way to consent to or opt out of the collection of satellite data or the development of SatML models that cover their land or community}.

SatML is characterized by a ``duality'' of potential impacts.
SatML technologies that can aid efforts in environmental and social justice by providing evidence~\cite{ovienmhada2023satellite,sefala2021constructing} can also be used to ``data-fy'' unjust practices like redlining, as \citet{sefala2021constructing} note.
Systems that forecast crop yields can improve crop insurance access~\cite{nakalembe2023considerations}, but inaccurate yield predictions can worsen outcomes for farmers~\cite{benami2021can}.
SatML models used for damage assessment after natural disasters also have military applications.

Existing efforts from responsible and ethical ML need significant adaptation to address the distinct concerns of SatML~\cite{kochupillai2022earth}.
For example, a model card template~\cite{mitchell2019model} could be specialized for SatML models to highlight potential for disparate performance across geographies~\cite{aiken2023fairness}, delineate boundaries of intended use~\cite{meyer2022machine}, and encourage use-contextualized model evaluations (\S\ref{sec:evaluation}).

However, we need to do more than adapt existing efforts.
The SatML community must collaborate with stakeholders about the trade-off between the risks and benefits of SatML technologies and understand how decisions made using SatML-generated maps can impact policy~\cite{sandker2021importance} and communities~\cite{szantoi2020addressing}.
As one step toward this, SatML efforts can integrate CARE principles for indigenous data governance~\cite{carroll2020care} and account for the unique context of satellite data~\cite{dogan2023you}. 
%

\section{SatML warrants specialized ML methods}
\label{sec:tailored}
Since satellite data can often be easily formatted as 3-channel images (e.g., by removing all but 3 spectral channels), many existing computer vision models can be applied ``out of the box'' to satellite data.
While this ``lift and shift'' strategy can 
facilitate rapid exploration of research ideas and failure modes, it is also constraining, as existing ML approaches are ill-fitted to many of the distinctions of satellite data (\S\ref{sec:satml_distinct_challenges}).
%


On the other hand, the SatML characteristics outlined in \S\ref{sec:satml_distinct_challenges} can drive novel ML algorithms, architectures, and models.
In contrast to the mainstream ``lift and shift'' approach, the following examples evidence how emerging SatML
approaches designed explicitly for the satellite data modality 
have increased performance or real-world applicability over traditional ML approaches.
\textit{We argue that the ML community must pursue and catalyze more such research that is designed for the needs of SatML}, rather than being content with methods that work for some tasks despite those needs. 

\subsection{Learning strategies} 
\label{sec:learning_strategies}

ImageNet pre-trained models are popular for transfer learning tasks. However, ImageNet only contains 3-channel (RGB) natural images. 
Naively adapting ImageNet pre-trained models for satellite data---for example, by using only the RGB channels of a multispectral image---is a clear example of ``lifting and shifting'' methods designed for natural images to satellite data. 
While ImageNet models have worked well for some SatML tasks after fine-tuning~\cite{lacoste2023geo}, they are much worse than random initialization for others~\cite{wang2022empirical}. 
\citet{bastani2023satlaspretrain} showed that pre-training supervised models with satellite datasets improved performance by as much as 18\% over ImageNet pre-training.

Representation learning methods designed specifically for SatML have also outperformed ``lift and shift'' approaches. Feature extraction using random convolutional features \cite{rolf2021generalizable,corley2023revisiting} and random pixel sets \cite{garnot2020satellite} are strong baselines for many SatML tasks, but generally are not for natural images. \citet{russwurm2024meta} proposed an adaptive meta-learning model designed for satellite data with a variable number of spectral channels and output classes. \citet{tseng2022timl} incorporated task-specific metadata like geographic region in the meta-learning process.

Self-supervised models pre-trained with Landsat or Sentinel-2 multispectral imagery have outperformed ImageNet weights in diverse SatML tasks~\cite{wang2023ssl4eo,stewart2023ssl4eo}. 
SSL models customized to the unique characteristics of satellite data such as the importance of scale~\cite{reed2022scale}, temporal dimension~\cite{tseng2023lightweight,manas2021seasonal}, and availability of multiple sensors~\cite{fuller2023croma} have outperformed approaches designed for natural images.
\citet{cong2022satmae} proposed a specialized encoding for groups of spectral bands to preserve relevance between individual spectral channels. 
\citet{ayush2021geography} added an auxiliary geolocation prediction task to a similar contrastive pre-training setup. \citet{tseng2023lightweight} and \citet{hackstein2024exploring} showed that, in a masked autoencoder framework, structured masking strategies outperformed conventional random masking when evaluated on downstream tasks.

\subsection{Model architectures} 
\label{sec:architectures}
To advance progress on SatML tasks, ML researchers need to re-consider the explicit and inductive biases present in popular deep learning model architectures.

For example, common deep learning architectures designed for natural image benchmarks are translation equivariant, but not rotation equivariant. 
This design choice stems from the fact that orientation in natural images is often important for preserving object context---predicting the ``correct'' orientation of rotated images is even an effective pretext task for pre-training~\cite{gidaris2018unsupervised}.
In contrast, objects in overhead satellite images do not have a ``natural'' orientation.
Considering the special characteristics of satellite imagery, \citet{marcos2018land} showed that rotation equivariant models improve performance in SatML land cover mapping tasks. \citet{van2018you} proposed specialized augmentations in satellite data for rotation invariance, as well as an architecture specialized to detect small, dense clusters of objects common in high-resolution satellite images (e.g., cars in parking lots).

Modern semantic segmentation models such as U-Net~\cite{ronneberger2015u} or DeepLabV3+~\cite{chen2018encoder} have large theoretical receptive fields, whereas some SatML tasks, such as land cover mapping, only require small receptive fields (if you assume land cover at a point is a function of the local spectral responses and textures at that point). \citet{brown2022dynamic} leveraged this constraint and other satellite data properties to achieve state-of-the-art land cover segmentation results with an architecture nearly $100\times$ smaller than standard segmentation architectures.

While most computer vision research uses datasets captured by a single sensor type (RGB photographs), satellite data is marked by diversity in sensors. Data fusion research ~\cite{steinhausen2018combining,van2018synergistic,whyte2018new,denize2018evaluation,tseng2023lightweight,babaeian2021estimation} has shown that models designed to leverage data from multiple complementary sensors can substantially outperform models trained with data from a single sensor.


\subsection{Explicitly modeling domain context}
Specialized SatML methods can encode geographic context and constraints. Spatial and spatio-temporal structures are omnipresent in satellite data and can inform model design and training procedures, or may provide inductive biases. 

For example, \citet{klemmer2022spategan} and \citet{klemmer2021sxl} incorporated autocorrelation measures into the model loss 
which improved performance in predictive and generative modeling of geospatial patterns. 
\citet{jean2019tile2vec} used inspiration from Tobler's first law of geography, that ``everything is related to everything else, but nearby things are related more'', to design an unsupervised triplet loss based on geographic distance between samples. 
\citet{mac2019presence} and \citet{mai2023csp} directly incorporated geographic priors into image classification models by conditioning model outputs on image locations.

Objects in temporal sequences of satellite images typically remain in a fixed position but change in appearance over time (unlike many videos). \citet{robinson2021temporal} leveraged the fact that many human-made structures do not change geographic position once constructed to design a parameter-free change point detection method for satellite time series.

\section{Satellite data enriches ML research}
\label{sec:satml_contributes_to_coreml}

By positioning satellite data as a distinct modality alongside other popular modalities like images and text, the entire field stands to make advances in the robustness, scalability, and flexibility of ML methods.
The geographic grounding, temporal regularity, global coverage, and sensor diversity of satellite data can inspire the creation of new datasets, experiments, and research directions, and catalyze progress on long-standing research questions.
As evidence, we highlight several ``core ML'' research areas that satellite data can enrich.

\subsection{Distribution shift}
ML models that are reliable, fair, and safe for real-world use must perform well across a variety of unseen distributions. Previous work in computer vision showed that ML models are highly sensitive to distribution shifts, even those that may be subtle or undetectable by humans~\cite{taori2020}.

\begin{figure}[t]
    \centering
    \includegraphics[width=\linewidth]{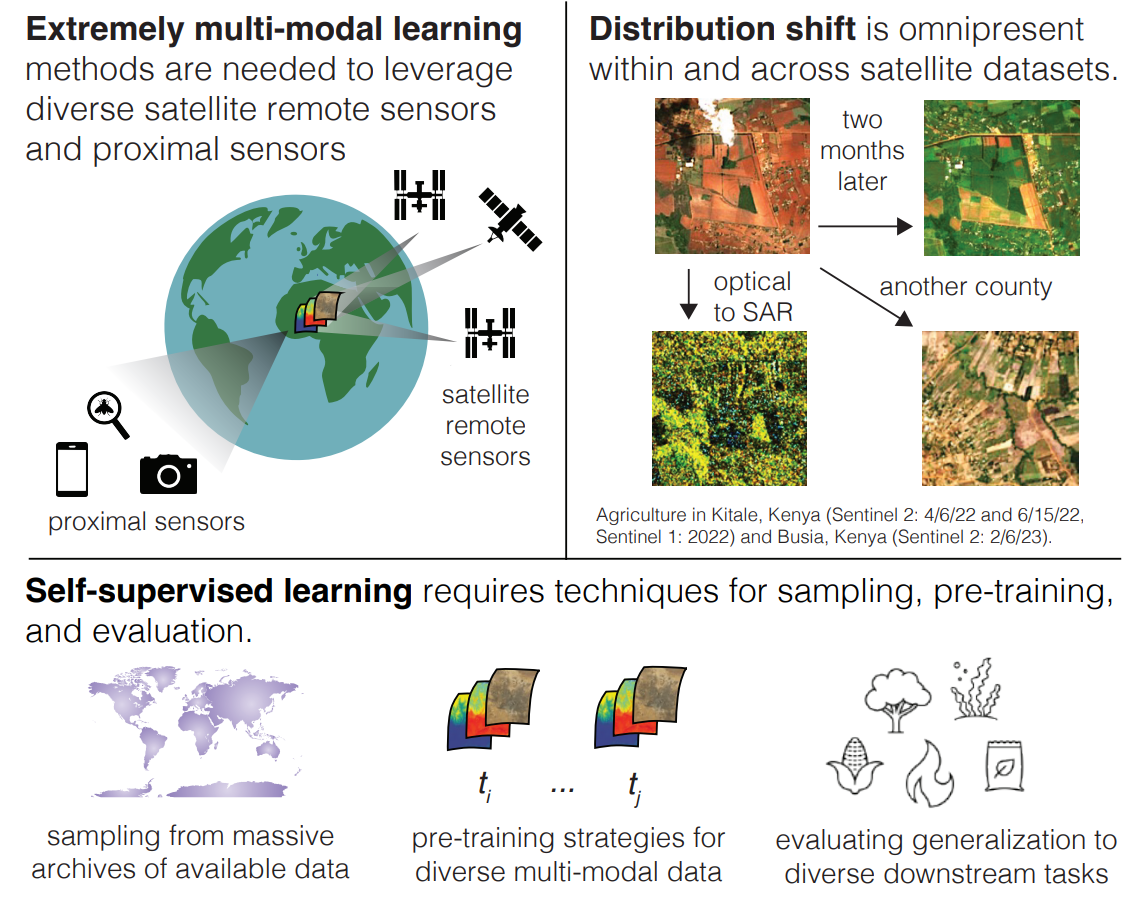}
    \caption{\textbf{SatML can enrich many research areas in ML}, e.g., multi-modal, self-supervised, and distributionally robust learning.}
    \label{fig:enrichingml}
\end{figure}

\textit{Distribution shifts are inevitable and pervasive in SatML contexts} (\cref{fig:enrichingml}). Covariate shifts occur across time, space, spectrum, and scale (\S \ref{subsec:sat_data_differences}). Concept drift occurs as environmental and social phenomena shift dynamically over time. 
Previous work suggests that ML models may generalize worse for satellite data than natural images. 
\citet{benson2020assessing} show that the accuracy of models trained to detect post-disaster building damage in satellite images dropped by as much as 30\% when evaluated on a new disaster site. Field delineation models transfer poorly across countries due to differences in farming practices, crop types, and growing seasons
\cite{kerner2023multi}.

The different wavelengths, temporal cadences, and spatial resolutions of satellite data (\S\ref{subsec:sat_data_differences}) have driven research in domain adaptation~\cite{tuia2016domain,song2019domain}.
Studying distribution shifts in satellite data  provides immense value for SatML use cases and increase the real-world usability of models. It can also enrich ML research by providing new datasets, case studies, and contexts for studying distribution shift and distributional robustness. For instance, the long historical archive of satellite data with global coverage provides a rich data source for studying distribution shift.

\subsection{Self-supervised learning and evaluation}
Self-supervised learning (SSL) is especially important for SatML, where unlabeled data abounds but labeled data is scarce and difficult to obtain (\S\ref{subsec:sat_data_differences}).
In \S\ref{sec:learning_strategies}, we gave examples of effective SSL strategies designed specifically for satellite data. In this section, we discuss how the unique context of SatML can enrich ML research on self-supervised learning.

%

\emph{The nature of satellite data motivates new SSL research questions} (\cref{fig:enrichingml}).
Since the availability, sensor types, and fidelity of satellite observations can vary greatly (\S\ref{subsec:sat_data_differences}), SatML is an excellent testbed for studying how to extract representations from complex, real-world data.
The diverse applications of SatML motivate SSL models capable of generalizing to a wide range of tasks~\cite{lacoste2023geo,yeh2021sustainbench,klemmer2023satclip}.
The sheer scale of satellite data raises questions about how to sub-sample data effectively for pre-training, 
and how data complexity relates to model capacity.

Beyond training, there has been limited research on approaches for \textit{evaluation} without labels. In SatML settings, it is rarely (if ever) feasible to acquire new labeled datasets for every new deployment setting (i.e., for new every year, for target, and region a model is applied in). This practical need adds urgency, real-world relevance, and new challenges to the emerging field of unsupervised, self-supervised, and semi-supervised evaluation~\cite{guillory2021predicting,deng2021labels,yu2022predicting,baek2022agreement}.

\subsection{Multi-modal learning}
Piecing together the most complete view of the Earth requires leveraging multiple data sources.
Satellite platforms collect and geographically align a variety of observations, including traditional cameras, thermal images, radar or synthetic aperture radar, and Lidar (\cref{fig:multi-sat-example}). 
Derived products like digital elevation models (DEMs) and soil maps~\cite{poggio2021soilgrids} can provide a semantic view of ground conditions. Despite this wealth of available sensors, few SatML methods today leverage inputs from more than 1-2 sensors (\S\ref{sec:architectures}) -- this is a promising and necessary direction for future ML research.

\emph{Satellite data brings rich, real-world challenges for multi-modal ML.} 
For example, fusing high-resolution imagery with low-resolution and partially-observed derived products requires 
multi-modal learning with coarse or weak observation modalities~\cite{rolf2022resolving,li2022deep}. Datasets such as SEN12MS~\cite{sen12ms} or CropHarvest~\cite{tseng2021cropharvest} that pair labels with observations from multiple satellites can serve as benchmarks for new algorithms.
There are also new, untapped opportunities for multi-modal learning combining raster (image-like) and vector (polygon-like) datasets~\cite{cepeda2023geoclip} that would enrich the scope of multi-modal learning studied by ML researchers today.

Satellite data can complement learning from other modalities including text~\cite{uzkent2019learning}, natural images~\cite{zhai2017predicting,dhakal2023sat2cap}, sound~\cite{khanal2023learning}, and other on-the-ground sensors (``proximal sensors'') by providing additional supervision or training signal~\cite{tuia2021toward,lefevre2017toward}.
The SatML task of cross-view matching also poses a unique multi-modal learning challenge in which the goal is to match ground-level images with their location in corresponding satellite images~\cite{hu2018cvm}.


\subsection{Positional encodings}
Positional encodings enable models to learn short- or long-distance interactions in naturally hierarchical data such as text (sequences) or images. They are an integral component in the success achieved by Transformer architectures~\cite{vaswani2017attention}. Satellite data contain additional spatial and temporal hierarchies and are distributed in a non-Euclidean, spherical space.

\emph{Encoding the spatio-temporal position of satellite data in neural network architectures is a distinct research challenge.} In recent years, studies on longitude/latitude location encoders for neural networks proposed sinusoidal~\cite{mac2019presence, mai2020space2vec} and spherical harmonics~\cite{rußwurm2023sh} representations. Spatio-temporal location encoders or location encoder pretraining are promising directions for future research. Future work could also investigate encoding additional satellite metadata, such as spatial resolution or sensor angles.


\subsection{Human-in-the-loop and active learning}
Since labeled data is generally rare but unlabeled data is abundant in SatML, 
iterating through phases of selective labeling and model retraining can be a promising approach~\cite{ortiz2022artificial,desai2022active,robinson2020human}. \textit{The unique nature of SatML raises research questions for human-in-the-loop and active learning} such as how to scale existing active learning query methods to petabyte scale unlabeled datasets. It also brings new constraints: e.g., some geographic areas are unsafe to visit, or more expensive to get to.

\section{Synthesis and Discussion Topics}
\label{sec:conclusion}
At first glance, SatML may seem like a straightforward application area of ML or computer vision. We argue that this perspective greatly undervalues the distinct challenges and potential of SatML research and that we must recognize satellite data as a distinct modality for ML research. 
However, much is at stake in intensifying SatML research without community discussion and consensus on how we will orient and measure research progress. 
%
Application without understanding---of the data modality, the downstream tasks, or the stakeholders---will not result in real-world impact.

Recognizing and approaching satellite data as a distinct modality in ML will help facilitate focused and responsible exploration in SatML and ML at large.
%
%
ML researchers aiming to innovate in the sub-field of SatML must address the unique nature of satellite data and SatML problem settings (\S\ref{sec:satml_distinct_challenges}) to design approaches that are purpose-driven, efficient, and impactful (\S\ref{sec:tailored}). 
For researchers who work primarily in other areas of ML, SatML data and contexts can amplify research impact and open opportunities for synergistic developments bridging theoretical and applied ML (\S\ref{sec:satml_contributes_to_coreml}).

The ML community has benefited from developing specialized approaches for distinct data modalities. For example, the convolutional neural network architecture arose from the need to detect objects with translation and illumination invariance in natural images~\cite{krizhevsky2012imagenet}. The Transformer architecture arose from the need to capture patterns and dependencies in  the language modality~\cite{vaswani2017attention}. The widely-used U-Net architecture was inspired by the needs of biomedical data~\cite{ronneberger2015u}.
We have not yet discovered what is possible for satellite data.


In addition to addressing the research challenges and opportunities discussed throughout this paper, realizing the full potential of SatML will require substantial shifts in the status quo for the wider ML community. 
Towards this, we pose the following discussion points and action items for the community:

\paragraph{How will we prioritize and collaborate on the most important SatML challenges?}
SatML has the potential to transform many research areas -- both within and outside ML -- with different norms and institutions. 
Organizing and supporting researchers from these different communities requires coordinated action.

\begin{figure}[t!]
    \centering
    \includegraphics[width=\linewidth]{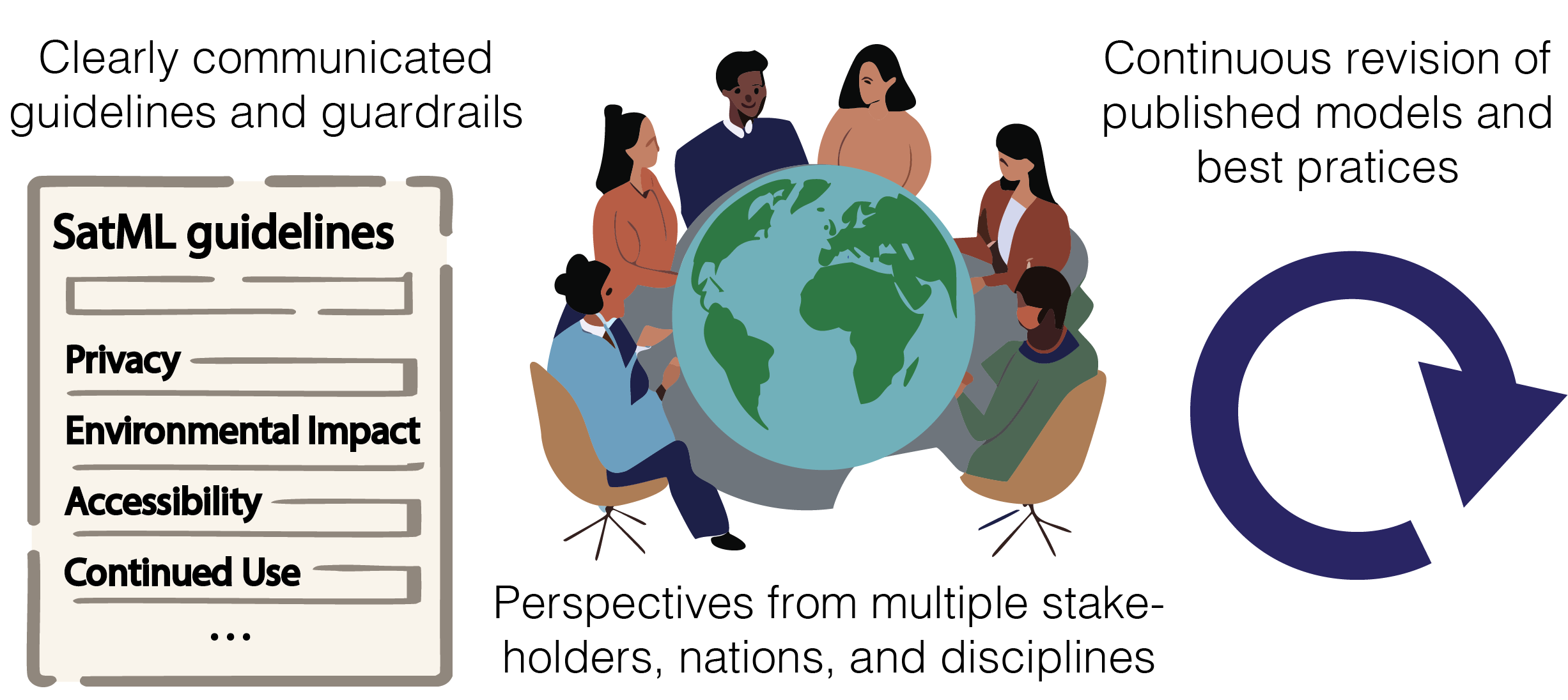}
    \caption{\textbf{Developing priorities and guidelines for SatML research} connects many communities, stakeholders, and disciplines.
    }
    \label{fig:ethics}
\end{figure}

\emph{We must develop an inclusive SatML research community with common priorities and guardrails (\cref{fig:ethics}).} 
SatML must integrate perspectives from many communities and experts within ML and fields like remote sensing, geography, the geosciences, and social sciences.
Otherwise, we run the risk of reinventing results that other disciplines have established decades ago, and applying computationally intensive methods for no real value added~\cite{tuia2021toward}.

\emph{Disciplinary institutions must acknowledge SatML's place within the scope of  ML and computer science research.}
The key institutions that facilitate research progress (e.g., publishing venues and funding agencies) dictate in large part how SatML research will proceed.
With the help of the SatML community, publishing venues must establish what research fits
(i) ``core ML'' venues (e.g., ICML), (ii) venues in other disciplines (e.g., remote sensing journals), and (iii) cross-cutting venues (e.g., remote sensing workshops at core ML venues). 
Funding agencies must support individual SatML research agendas that advance research beyond the status quo, which in aggregate can influence discipline-level paradigm shifts. 

\paragraph{How will we align SatML research progress and real-world impact?}
Historically, ML progress has been catalyzed by benchmark datasets and challenges: where there is data, there is an opportunity to climb the leaderboard.
In contrast, SatML is often motivated by on-the-ground information needs (e.g., from an end-user or research partner who will use the produced model). 
For SatML, we must reconsider how we abstract core challenges into benchmarks and how we measure research progress and contributions.

\emph{SatML benchmarks must be designed to catalyze impact.}
Massive volumes and varieties of satellite datasets are available, however it is still unclear which of the existing benchmarks and metrics are the best proxy for impact.
Ideally, SatML benchmarks and metrics will (i) emphasize the challenges that require advances in core ML (\S\ref{sec:satml_contributes_to_coreml}) and (ii) realistically capture real-world data and deployment conditions (\S\ref{sec:satml_distinct_challenges}). 

\emph{We need to emphasize SatML research in the real world.}
Studies on deploying SatML in on-the-ground conditions will not propose a new algorithm or dataset, but their results can add critical knowledge about the performance or utility (or failure points) of SatML in real-world conditions.
We must synthesize findings from deployed SatML research to revisit our established practices, benchmarks, and research directions. 
We could explicitly value and catalyze such research by designing SatML-specific ``impact challenges'' or other community milestones to orient research toward long-term real-world impact~\cite{wagstaff2012machine}.
\paragraph{How will we ensure SatML progress benefits global and local communities?}
When choosing interesting SatML research directions, ML researchers must consider how their research can impact real communities and ecosystems.


\emph{We must value global and community-focused research.}
 Some of the biggest potential for SatML impact lies in low-income or under-resourced countries and communities. To benefit from SatML solutions, many end-users need efficient, low-resource ML models due to compute access, internet cost/speed, and other constraints~\cite{nakalembe2023considerations}. The ML community should value research contributions that make SatML technologies accessible to end-users globally by addressing these constraints ~\cite{rolf2021generalizable}, rather than scoping them as outside of the research domain.
%
%

\textit{We must re-think our open science practices.}
The standard practice of openly sharing data, code, and models has been fundamental to rapid progress in ML research. However, we must question this practice when data represent human communities and activities and models can be used for unintended purposes (\S\ref{sec:ethics}). We must create new ways of enabling research progress while minimizing harm and respecting communities, for example through new models of data stewardship~\cite{walter2021indigenous} and data sharing~\cite{sefala2021constructing}.

\section*{Acknowledgements}

We thank Marc Ru\ss wurm, Claire Monteleoni, Sherrie Wang, David Rolnick, Anthony Ortiz, Isaac Corley, Patrick Gray, Gabriel Tseng, Brice Ménard, and Max Simchowitz for their valuable feedback on the arguments supporting our position. We thank Mirali Purohit, Rahul Nair, Snehal Chaudhari, Aditya Mohan, and Ivan Zvonkov for feedback on writing and clarity.

\emph{Funding.} ER is supported by the Harvard Data Science Initiative, the Center for Research on Computation and Society at Harvard, and Microsoft.

\bibliography{satml-position}
\bibliographystyle{icml2024}

\newpage
\appendix
\onecolumn
\section{Related workshops} \label{appendix:workshop_list}

We are aware of 19 instances of workshops since 2018 that have been hosted at CS/ML conferences and that promote SatML research (Table \ref{tab:workshops}).

\begin{table}[h]
\centering
\resizebox{\columnwidth}{!}{%
\begin{tabular}{@{}llc@{}}
\toprule
\textbf{Workshop Name} & \textbf{Conference Name} & \multicolumn{1}{l}{\textbf{Years active}} \\ \midrule
DeepGlobe & CVPR & 2018 \\
EarthVision & CVPR & 2015, 2019-2024 \\
Object Tracking and Classification in and Beyond the Visible Spectrum & CVPR & 2004-2011 \\
Perception Beyond the Visible Spectrum & CVPR & 2012-2023 \\
Machine Vision for Earth Observation and Environment Monitoring & BMCV & 2023 \\
Complex Data Challenges in Earth Observation & CIKM, IJCAI & 2021, 2022 \\
GeoAI & SIGSPATIAL & 2019, 2021, 2023 \\
Machine Learning for Remote Sensing & ICLR & 2023-2024 \\ \bottomrule
\end{tabular}%
}
\caption{Workshops among CS/ML conferences that promote SatML related research.}
\label{tab:workshops}
\end{table}


\end{document}